# Time-aware topic identification in social media with pre-trained language models: A case study of electric vehicles

Byeongki Jeong[a], Janghyeok Yoon[b], Jaewoong Choi[*,c]

[a] Optimization & Analytics Office, SK Innovation, Seoul, Republic of Korea

[b] Department of Industrial Engineering, Konkuk University, Seoul, Republic of Korea

[c] Computational Science Research Center, Korea Institute of Science and Technology, Seoul, Republic of Korea

[*]Corresponding author: jwchoi95@kist.re.kr

**Abstract**

Recent extensively competitive business environment makes companies to keep their eyes on social media, as there is a growing recognition over customer languages (e.g., needs, interests, and complaints) as source of future opportunities. This research avenue analysing social media data has received much attention in academia, but their utilities are limited as most of methods provide retrospective results. Moreover, the increasing number of customer-generated contents and rapidly varying topics have made the necessity of time-aware topic evolution analyses. Recently, several researchers have showed the applicability of pre-trained semantic language models to social media as an input feature, but leaving limitations in understanding evolving topics. In this study, we propose a time-aware topic identification approach with pre-trained language models. The proposed approach consists of two stages: the dynamics-focused function for tracking time-varying topics with language models and the emergence-scoring function to examine future promising topics. Here we apply the proposed approach to reddit data on electric vehicles, and our findings highlight the feasibility of capturing emerging customer topics from voluminous social media in a time-aware manner.

# 1. Introduction

Recent years witnessed the increasing importance of understanding customer languages in competitive business environment (Xu, Wang, Li, & Haghighi, 2017), as the influence of customers affects not only other customers, but also new product planning and improvement, and competitive intelligence (Jeong, Yoon, & Lee, 2019). Thus, the ability of companies to collect, analyse and reflect the opinions of customers promptly has become indispensable. Especially, transforming customer language on needs, complaints or interest into business opportunities can guide companies to obtain competitive edges over their competitors. As the utility of big data of customer-generated data increases with the development of social media, companies are trying to keep their eyes on what customers says on their product, services, or brand in real-time. Several companies have constructed their own social media channels to detect and track the needs, or complaints of customers, as well as to reflect the expectations of potential customers in new product planning (Choi, Oh, Yoon, Lee, & Coh, 2020; Olanrewaju, Hossain, Whiteside, & Mercieca, 2020).

Also, in the academia, there have been many data-driven attempts to customer understanding, most of which are based on the application of text mining, deep learning, or network analyses in social media data (Choi, Yoon, Chung, Coh, & Lee, 2020). Among the various research questions, the one regarding which customer topics or needs are most important as business opportunities has been actively discussed. As companies cannot reflect all the customer needs into their business, the prioritisation of customer-stated topics is required in addition to the topic detection. In this research avenue, some pioneers have suggested an approach based on latent Dirichlet allocation and opportunity algorithm to identify topics and evaluate their opportunity level (Jeong, et al., 2019). Similarly, there have been novel methods based on network structure of customer-stated terms (Misuraca, Scepi, & Spano, 2020), pre-trained language models (Choi, Oh, et al., 2020; Ozcan, Suloglu, Sakar, & Chatufale, 2021), or topic modelling (Ko, Jeong, Choi, & Yoon, 2017; Nolasco & Oliveira, 2019) to identify the potential topics from voluminous customer-generated data. These studies have significant contributions by providing data-driven ways of utilising various social media data for the sake of customer understanding. The systematic method and scientific results made it possible for subsequent researchers and companies to reproduce the methodology according to their environment.

However, as the volume of customer-generated contents increases in real-time, and the topics customer stated are time-varying (Chong & Chen, 2010; Yin, et al., 2016), prior methods to customer understanding showed the limited utility in practice. The focus of previous data-driven methods lies in identifying which customer-stated topics exist at the time of analyses, which makes their utility limited. That is, such methods are applicable to static data structures, thereby providing cross-sectional results for the given dataset. This mechanism allows companies or researchers, users of the methodology, to only retrospectively analyse past topics identified from accumulated consumer-generated data. Additionally, previous studies have not excluded labour-intensive processes in applying data analyses techniques to consumer-generated data, particularly in text pre-processing, filtering, and even interpretation. In a rapidly changing business environment, these points can act as a fatal disadvantage for companies that need to understand customer-stated topics and create business opportunities more agile than competitors.

In response to these issues, this study proposes a new method to identify the history of customer-stated topics and anticipate the futures of topics using emergence scoring, thereby fully automating the customer language understanding process. The main goal can be summarised as understanding the past and future of the topics and their possibilities as business opportunities. To this end, we design two functions of tracking the history of topics such as birth, growth, decay, and death, and anticipating the future of topics with quantitative emergence scores. This automatic approach does not require expert intervention and is robust against changes, updates, and additions of the initial dataset. We applied this approach to the dataset of social media to test the detection of customer-stated topics related to Tesla motors Inc.

The paper is organised as follows: we review prior studies on data-driven ways of customer language understanding, discussing their contributions and limitations. The third section presents the process of the proposed approach: we describe the overall process by step, including the functions we designed to analyse the past and future of customer-stated topics. The fourth section presents the case study on the Reddit dataset, social media community to detect emerging topics regarding Tesla motors Inc. The last section discusses the implications of results as business opportunities and concludes with the limitations and future works of this study.

## 2. Background on customer topic analyses

This study suggests a data-driven approach to understanding customer language by analysing the topics customer stated in social media. The identification of customer-stated topics has received much attention in recent years, especially in business opportunity-related research avenues (Choi, Yoon, et al., 2020). Therefore, we investigated prior studies which addressed similar research questions on customer-stated topics, by comparing the ways of defining, identifying, and evaluating customer-stated topics and discussing their contributions and limitations.

Going back to the origins of research questions about consumer language, in the early days, customer needs studies have mainly focused on technical requirements for products that could be addressed by engineering (Griffin, 2004). However, in recent years, researchers have also turned their attention to analysing and predicting customer behaviour including customer satisfaction (Udo, Bagchi, & Kirs, 2010), customer engagement (Kang, Lu, Guo, & Li, 2021), and customer churn (Haenlein, 2013), or uncovering consumer needs that can generate new ideas or opportunities for business (Jeong, et al., 2019; Misuraca, et al., 2020). From a methodological point of view, beyond statistical analyses of data such as focus group interviews (FGIs) and surveys, recently, large amounts of consumer-generated data (e.g., social network service, blogs, comments) have been covered by network analyses, topic modelling, and machine learning. Numerous new technologies are being applied to analyse consumer needs and opinions. These data-driven methodologies require a customer-generated data collection process and the previous methods used to collect customer language can be divided into two groups: direct and indirect approaches.

First, direct approaches such as FGI and surveys involve directly interacting with customers to collect customer-generated data (Wang & Tseng, 2013). FGI is a traditional and intuitive way to collect customer voices face-to-face. This is mainly conducted in the form of a group discussion where selected customers can freely express their various opinions about products, services, or brands, and managers or in-house experts should observe the discussion and gather customer perceptions, opinions and preferences for companies' values. Surveys are a way to get customer responses to prepared questions, and this traditional method has the advantage of being able to easily obtain data suitable for the companies. However, through the survey method, companies communicate with consumers in a passive way, and this method is costly and time-consuming above all else. Also, in this method, it is difficult to identify potential interests that respondents are not aware of other than the questions. This leads to a disadvantage in that it is not possible to collect potential needs or diverse opinions in the current business environment in which consumers and companies actively interact, and various opinions of consumers are used as a source of product innovation or business innovation.

However, due to the availability and accessibility of large amounts of data through social media, an indirect approach is attracting attention. The key to this approach is to collect customers' opinions through access to posts voluntarily created by customers, rather than directly listening to customers' voices (Olanrewaju, et al., 2020). With the development of the Internet and social media, a large amount of customer-generated data is accumulated in real time, and this data has a high analyses value because it affects not only the company's business but also other customers' product life cycle. In addition, as data analyses techniques that efficiently handle and analyse large amounts of data have been developed, various forms of social media data and analyses techniques have been actively used in academia along with indirect data collection methods. There have been many attempts to identify customer-stated topics in online review data for a product of interests (Roh, Jeong, Jang, & Yoon, 2019; Seo, Seo, Jang, Jeong, & Kang, 2020; Zhou, Jianxin Jiao, & Linsey, 2015). Zhou, et al. (2015) collected product reviews from the e-commerce platform, Amazon and used text mining and association rule mining to identify customer-stated topics, which were used to discover potential customer needs by applying analogical reasoning. Roh, et al. (2019) used sentiment analyses on product review websites to identify topics mentioned by customers. Relationships between identified subjects and patents were defined using SLM, and technology opportunities were discovered using those relationships. Seo, et al. (2020) attempted to identify anomalous customer opinions using SLM and anomaly detection methods utilising online vehicle reviews. The anomalous opinions identified in the study were visualised and used to understand customer needs for their vehicles.

Although several indirect approaches to identify customer-stated topics from online review data have been studied, this data only includes comments from customers who have already experienced a product or service. In other words, online product review data is difficult to cover every aspect of a topic mentioned by a customer, such as the interests or expectations of potential customers. This data mainly consists of information about individual products, which makes it to identify the subject matter of customer statements for the entire business or product line difficult. Consequently, other researchers preferred to use social networking services, online communities, and online forum data—which are more comprehensive than online review data—in their studies. Li, Xie, Jiang, Zhou, and Huang (2019) tried to connect customer topics and related technologies after extracting topics by applying topic modelling to Twitter data. Jeong, et al. (2019) used topic modelling on Reddit data to identify customer statement topics related to a specific product and quantitatively evaluate satisfaction and importance. They tried to identify topics that can become product opportunities and provide information on product improvement/innovation. Some researcher showed an significant example of utilising Facebook data to identify customer topics and profile user types based on their liking patterns (van Dam & Van De Velden, 2015). Choi, Oh, et al. (2020) applied an event detection and tracking method to online forum data, thereby identifying topics mentioned by customers for products and monitor changes at the keyword level.

The above-mentioned studies attempted to identify a broad range of customer statement topics, including the voices of potential customers and the opinions of experienced customers. However, most of these studies are based on cross-sectional analyses of customers from a static perspective, making it difficult to detect time-varying customer-stated topics in an effective way. As mentioned earlier, because customer-stated topics are dynamic by nature, identifying changes in topics over time is an important factor in this field of research (Chong & Chen, 2010; Yin, et al., 2016). Although there have been several attempts to detect these changes (Alam, Ryu, & Lee, 2017; Choi, Oh, et al., 2020; Li, et al., 2019; Lu, Tan, & Li, 2019), these studies lack consideration of the following issues: First, in most studies, the analyses of newly imported data influences the results of existing analyses. In other words, an approach that analyses topic modelling results with time slices or applies event detection algorithms requires integrating existing and new data and updating experimental results. Meanwhile, in this study, we propose a flexible dynamics-focused function that can identify customer-stated topics independently for each time slice and connect customer-stated topics in different time slices. Second, unlike previous studies that provide retrospective results by observing topics stated by customers using historical data, this study introduces an emergence scoring function that evaluates the emergence score of each topic and predicts the near future. Through the proposed approach, not only the history in customer-stated topics but also emerging customer-stated topics can be identified, which can guide firms to achieve competitive edges by pre-emptively responding to essential topics in the near future. The differences between previous studies and present research are summarised in Table 1.

Table 1. Comparison of previous studies and present research

| Factor | Static customer-stated topic analyses methods | Dynamic customer-stated topic analyses methods | The proposed approach |
|---|---|---|---|
| Focus | Cross-sectional results of customer-stated topics at the analyses time point | Time-varying changes of customer-stated topics during the analyses period | Dynamic changes of customer-stated topics from the past to near-future |
| Data | Online product reviews from e-commerce, online forum, and online questionnaires | Social networking service, online communities, and online forum data | Social media data including online forum data |
| Purpose | Identification of latent customer needs, product opportunities, technology opportunities or unusual customer responses | Detection of time-evolving customer-centric product opportunities or trend of emerging technologies | Anticipating the near-future of topics; Tracking the time-varying changes of customer-stated topics |
| Methodology | Topic modelling, anomaly detection, naïve Bayesian classifier, support vector machines, and case-based reasoning | Topic modelling, event detection and tracking algorithm, and sentiment analyses | Pre-trained SLMs, clustering, TOPSIS and sentiment analyses |
| Results and implications | Customer-stated topics related to product opportunities, unusual topics, latent topics are identified and evaluated at the analyses time point | The time-varying changes of customer-stated topics related to product opportunities, or emerging technologies are detected during the analyses period | The dynamics of customer-stated topics regarding products and the near-future of topics with quantitative emergence scores |

## 3. The proposed approach

3.1. Overview

The proposed approach is illustrated in Fig. 1. As this study utilises social media data as main source of customer language, the first step is collecting and pre-processing the customer-generated data. Most of customer-generated data is unstructured and thus requires a series of pre-processing for facilitating the subsequent analyses. The second step aims to detect customer-stated topics by time slice. At the heart of this step is the identification of topics by applying a pre-trained language model and unsupervised clustering methods to customers' keywords. The third step is the linking step. Here, linking aims to discover potential association relationship between topics by time slice, which enables to identify the history of topics. The final and most important step is the emergence scoring step. The objective of this step is to find the topic of future importance among the current topics, by measuring emergence score based on novelty, growth, community and coherence.

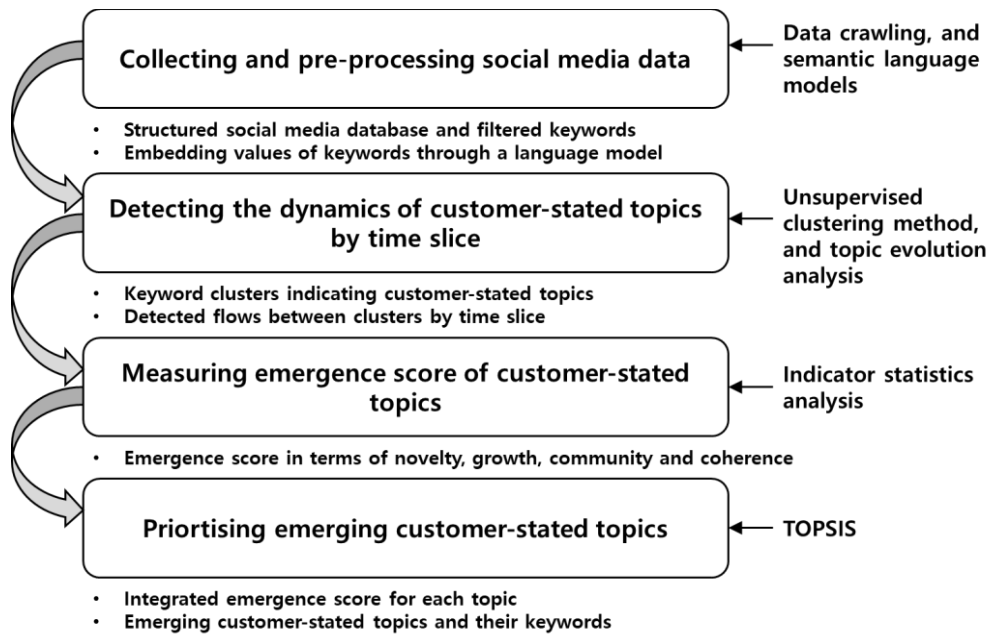

**Figure 1.** The overall procedure of the proposed approach

3.2. Data collection and pre-processing

Social media has become the most accessible and flexible windows where customers freely share, find, read and post their opinions, feelings, or experience on products, companies, services or brands. With the advance of big data processing technologies, many researchers can access social media data and analysing the customer-generated data for customer language understanding (Choi, Yoon, et al., 2020). Also, the proposed approach starts with utilising social media data related to a subject of interest such as a brand, product or company. The collection of social media data can be implemented by web scrapping or application programming interface (API). Using the API services is one of the efficient and precise ways to collect big data from social media platforms, as listed in Table 2. However, when a social media platform does not provide its own API, web scrapping would be the only systematic way to data collection. Here, python open library such as selenium can be utilised. Of course, there are often publicly available datasets from online data warehouses such as Google Big Query or Kaggle, but the results are not collected as desired for analyses. Finally, the output of data collection includes text data generated by customers with its creation time information.

**Table 2.** Social media platforms and their API references

| Social media platforms | API references |
| --- | --- |
| Twitter | https://developer.twitter.com/en/docs |
| Facebook | https://developers.facebook.com/tools/ |
| Reddit | https://www.reddit.com/dev/api/ |
| Instagram | https://www.instagram.com/developer/ |
| Yelp | https://www.yelp.com/fusion |
| Amazon | https://docs.aws.amazon.com/AWSECommerceService/latest/DG/ |
| Flickr | https://www.flickr.com/services/api/ |
| TripAdvisor | https://developer-tripadvisor.com/content-api/ |
| Qzone | https://support.qzone.gr/api |

As text data generated by customers is often unstructured, which makes it difficult to use for computational analyses in its original state. Therefore, applying text mining techniques to social media data requires the data pre-processing as a prerequisite process, where customer-generated data is transformed into an analytic form. The purposes of text pre-processing in this study can be summarised into two folds; One is to reduce noisy data for increasing the interpretability of intermediate or final outcomes, and the other is to transform unstructured text data into a structured form that can be utilised in the next topic evolution analyses step. The pre-processing can be conducted in the order of keyword extraction, keyword filtering, and keyword vectorization, as shown in Table 3. Here, the keyword means the smallest unit of analyses that can grasp the meaning of the social media text generated by customers, and may be a word or a noun phrase, a noun chunk. Noun phrases and subject-action-object structures are commonly used to recognise keywords, which can be extracted using open natural language processing libraries such as spaCy, StanfordNLP, or IBM Watson NLU. Although these tools perform acceptable in keyword extraction, most of them are domain-neutral, so their naive use requires some filtering of the extraction results. That is, among the extracted keywords, it is necessary to filter out keywords that are difficult to use in analyses as a consumer topic, that is, noise keywords. Generally, there are rule-based, stopwords-based, and frequency-based approaches to keyword filtering. The rule-based and stopword-based approaches use predefined rules and stopwords dictionary respectively. A rule-based approach removes keywords that are difficult to understand on their own, such as punctuation and pronouns, where POS tagging results can also be utilised. Also, by setting a limit on word length as a rule, you can exclude overly long keywords that are likely to be misspelled, or filter out keywords that are too short to understand their meaning. Finally, frequency-based approaches use metrics on a per-keyword basis, such as term frequency and document frequency. Here, it is assumed that common keywords appearing in most documents have low utility value in analyses. Also, keywords with too low a document frequency may have a negligible impact, or may be replaced by other high frequency keywords, so they may also be excluded.

**Table 3.** A systematic procedure for data pre-processing

| Step | Methods | Description |
| --- | --- | --- |
| 1. Extracting keywords from documents | Keyword extractor or tokenizer of open NLP library | - Utilise Noun chunks, keywords, or noun phrases as units of analyses |
| 2. Filtering noise keywords | Rule-based approach | - Exclude punctuation, and pronoun<br><br>- Remove outliers based on the length distribution of keywords |
| | Stopwords-based approach | - Use well-defined stop-word dictionaries provided by open libraries |
| | Frequency-based approach | - Remove outlier keywords using document frequency, keyword frequency, or term frequency-inverse document |

|   |   |   |   |
|---|---|---|---|
|   |   |   | frequency |
| 3. Vectorizing filtered keywords | Pre-trained language models such as BERT, RoBERTa, T5 | - Represent the selected keywords as high-dimensional dense embedding vectors |   |

After filtering out noise keywords, it is necessary to transform the selected keyword set $V$ into a form usable for the following computational analyses, i.e., vectorization. Vectorization of text data is the most prerequisite for text mining and natural language processing, as most computational algorithms can generally be referred to as a mapping process between input vectors and output vectors. When it comes to text vectorization, the bag-of-words (BoW) method is the most traditional and basic approach that assumes each word as a dimension. However, the BoW approach suffers from the curse of dimensionality because, in general, the number of words used in the analyses increases as the set of documents increases. This creates catastrophic problems in both the time and spatial complexity of algorithmic computation. Also, because BoW recognises words with the same or similar meaning but different expressions in different dimensions, semantic information cannot be used for analyses. To solve this problem, recent studies have used SLMs such as BERT, RoBERTa, and T5 to construct distributed representations of fixed-dimensional terms. For more efficient and accurate customer language understanding, Chen, Zhuo, and Wang (2019) used the BERT for customer intent classification, Day and Lin (2017) utilised the Word2vec for customer sentiment analyses, and Zhao, et al. (2015) designed the Paragraph2vec for product recommendation. Likewise, the text vectorization of the proposed approach is also depending on the strength of SLMs. The dimension size of the vectorized key terms, $V_d$, should be $\sum_t |V_t| \times d$, where $V_t$ is a set of key terms at time $t$ and $d$ is a dimension of an SLM. $v_{t,i}$ represents an $i^{th}$ term in $V_t$, and each $V_t$ can have an intersection.

3.3. Customer-stated topic evolution analyses system

We determined to adopt unsupervised clustering methods to the vectorised keywords by time slice, thereby enabling the proposed approach to be developed into a systematic customer-stated topic evolution analyses system. As high dimensional vectors of keywords encoded by pre-trained language models represent semantic meanings well, the application of unsupervised clustering to the keyword set can provide the group of keywords with similar meanings. Unlike the probability estimation-based method, this text mining approach provides a group of keywords with close semantic distance, enabling intuitive interpretation. We believe that once a keyword cluster for a topic is identified, the meaning of representative keywords represents the opinion of consumers on that topic. This study proceeds with further steps, assuming that all keyword clusters are sources of potential business opportunities as customer-stated topics. Also, unlike studies that provided cross-sectional analyses results, this approach detects topics by time slice and tracks the changes in topics. In this regard, we propose a method of linking detected topics by time slice. Link generation of topics follows a quantitative procedure that identifies only semantically related topic relationships.

3.3.1. Topic detection

The process of applying unsupervised clustering methods to a set of keyword vectors is illustrated in Algorithm 1. This study utilised k-means clustering, which is often used in analysing high dimensional vectors. Algorithm 1 finds the optimal number of clusters, $k$, for the embedded keyword vectors of each time slice $t$, $V_t$. We used the optimal number of clusters to group each of $V_t$. One of the important tasks in clustering is to find the optimal number of clusters. If the number of clusters is too large, very fine classification is possible, but it is difficult to find the overall flow. Conversely, if the number of clusters is too small, a broad search can be performed, but the specific structure is difficult to grasp. The optimal number of clusters can be determined through the elbow method based on the sum of squared errors within the cluster or the silhouette method based on the degree of cohesion within the cluster. The proposed approach uses gap statistics (Tibshirani, Walther, & Hastie, 2001) that are robust even for a high number of clusters, which presents a comparison of the overall within-cluster variation for *k* values different from the expected value in a null-referenced distribution of data using a Monte-Carlo method. Because each time slice *t* has its own value of $k$, the size of a set of clusters $C$ is $\sum_t k_t$. To interpret each customer-stated topic ($c_{t,i}$), representative keywords for each topic can be identified by calculating the distance between the constituent keywords and the centroid of the cluster. Here, we used the Euclidean distance, as in k-means clustering.

**Algorithm 1.** Detecting customer-stated topics by time slice

---

**Input:** embedded vectors $V = \{V_1, \ldots, V_T\}$, a maximum number of cluster $N$

**Output:** clusters $C = \{C_1, \ldots, C_T\}$ where $C_t = \{c_{t,1}, \ldots, c_{t,N}\}$ and $c_i \subset V_t$

  **Initialise** $C = \emptyset$

  **For** $V_t$ in V **do**

    **Initialise** k = 0

    **For** n = 2 to N **do**

      **Calculate** *GAP_STATISTICS* $g_n$

---

        **If** $g_n > k$ **then** k = n

    **End for**

    $C_t = K - MEANS(V_t, k)$

    **Insert** $C_t$ **into** C

**End for**

---

### 3.3.2. Topic linking

After the clustering was performed independently for each time slice, an additional process is required to understand the history of customer-stated topics. To this end, Algorithm 2 is designed to generate the relationship between the clusters in different time slices. In Algorithm 2, we used an intersect measure to determine the weight of relationship, which can be modified into the Jaccard coefficient or overlap coefficient according to the property of data. The outcomes of linking customer-stated topics between time slice can be represented as a weighted directed network, $G = (C, E)$, between the topics. As the network $G$ represents a dynamic flow between customer-stated topics over time, a visualisation such as a Sankey diagram can help identify the flow of topic changes. In addition, the relationship between topics can be interpreted by comparing the amount of backward and forward topics. For example, a relationship of one backward topic and one forward topic implies that persistent content emerges over time, and therefore is more likely to address important but unresolved issues or essential needs. The relationship in which multiple backward topics are transformed into one forward topic represents a flow in which slightly different needs are integrated into one opinion. This may include customer needs that were minor in the past but will be important in the near future. Conversely, one backward topic may be divided into several forward topics over time, which implies that customers' opinions and needs have diversified or the size of related detailed topics has increased. Finally, the relationship between multiple backward topics and forward topics is highly likely to be a marginal relationship just before a clear relationship is identified, so continuous monitoring is required.

**Algorithm 2.** Linking customer-stated topics between time slice

---

**Input:** clusters $C_t = \{c_{t,1}, \dots, c_{t,N}\}$, a threshold $\theta$ $(0 \leq \theta \leq 1)$

**Output:** weighted directed network $G = (C, E)$

---

```
Initialise G = ∅
For t = 1 to T − 1 do
  For c_i in C_t do
    For c_j in C_{t+1} do
      w_{i,j} = |c_i ∩ c_j|
      If |c_i ∩ c_j|/|c_i ∪ c_j| > θ then Insert w_{i,j} into E
    End for
  End for
End for
```

### 3.3.3. Emergence scoring

After the detection of customer-stated topics using the two algorithms, the topics should be investigated to identify which topics are valuable as opportunities. To this end, this step utilises the emergence scoring function to identify emerging customer-stated topics which are expected to be of greater importance in the near future. Although many researchers have discussed the characteristics of emerging topics in the fields of business, politics, medicine and management, the concept of emergence is still controversial as it possesses an ambiguous and uncertain character. Many researchers have suggested and quantified various proxies of emergence, among which novelty, growth, community, and coherence are common and frequent indicators (Garner, Carley, Porter, & Newman, 2017). Some researchers have argued that novelty and growth are the most key attributes of emergence (Rotolo, Hicks, & Martin, 2015), as the rapid growth of previously unobserved objects is likely to influence future landscapes. The novelty or newness of an analyte is generally measured by its heterogeneity with a comparator (other subjects), which is often used as a potential source of innovation or weak signals (Kim & Lee, 2017; Small, Boyack, & Klavans, 2014). In this study, the novelty of a topic is calculated based on its heterogeneity with other topics within the same time slice, which can indicate the degree of uniqueness and innovativeness of a topic mentioned by customers. Growth indicators are based on a quantitative assessment of the emergence of an analyte, assuming that those that emerge incrementally over a period of time are more likely to emerge in the near future than those that do not (Cozzens, et al., 2010). In addition, we use the size or influence of the set of parties (called players) associated with the analysed as an indicator of emergence, assuming that the emerging of the analysed is influenced by the activities or powers of the players involved. Several researchers have quantified community metrics based on citation relationships between components or networks between players. Finally, the coherence indicator (Alexander, Chase, Newman, Porter, & Roessner, 2012; Garner, et al., 2017), which indicates the consistency and momentum of the analyte, is utilised as an attribute of emergence. Some researchers argued that the unification of the types of objects related to the analyte shows coherence, and they interpreted the trend of decreasing the number of words related to topics as a signal of emergence (Rotolo, et al., 2015).

Based on the above-mentioned 4 indicators, this study defines measures of emergence properties $c_{t,i}$ for customer-stated topics of the last time slice $T$. First, to capture the novelty attribute, this approach uses the proportion of new keywords at time $T$ among all other keywords in the topic, which yields intuitive results and leverages existing results. In this study, if a topic consists of new keywords that did not exist before, it is likely that the topic is dealing with novel content and can satisfy the requirements of emergence. Second, this study uses the growth rate of the number of documents related to the keywords of a topic as a growth indicator because growth can be indicated by a quantitative increase in a topic. That is, a topic with an increasing number of related documents is likely to be frequently dealt with by many consumers, so it may contain content that will become increasingly important in the near future. Next, community attribute is typically expressed as the number of players for each topic, so we calculate it based on the average number of customers referencing each keyword in the topic. If keywords related to a topic are mentioned by many customers rather than a specific customer, the importance of the topic can be considered high. Finally, coherence is indicated by using semantic consistency of keywords belonging to a topic. Specifically, it computes the mean of cosine similarity between all pairs of semantic vectors in a topic. For example, since topics composed of keywords with similar meanings have high consistency, they are highly likely to contain common opinions of customers on specific needs or issues.

3.3.4. Topic prioritisation

In the previous step, quantitative immersion indicators were designed and calculated to identify emerging topics, but since the values and ranges of indicators are all different, the issue of ambiguity in priorities between topics remains. By design, novelty and coherence can have values between 0 and 1, growth can have values greater than 0, and community can have values greater than the inverse of the absolute value of $c_{t,i}$. Because these four emergence attributes have different scales, simply adding or multiplying them is not appropriate. In addition, since the importance of each attribute may be different by analyst or dataset, a method that reflects various importance criteria is required. Therefore, it is natural to include normalisation of values and weighting between indicators in the process of integrating indicators. To this end, we utilise the TOPSIS (Hwang & Yoon, 1981), a representative multiple-criteria decision making method for prioritising emerging customer-stated topics and developing an integrated emergence score. The assumption of TOPSIS is that the alternative which is closest to the positive-ideal solution and farthest from the negative-ideal solution is the best. To prioritise the alternatives, TOPSIS computes the geometric distance from both the positive and negative ideal solutions, as shown in Figure 2. Consequently, the four emergence indicators are aggregated into one indicator representing emergence via TOPSIS.

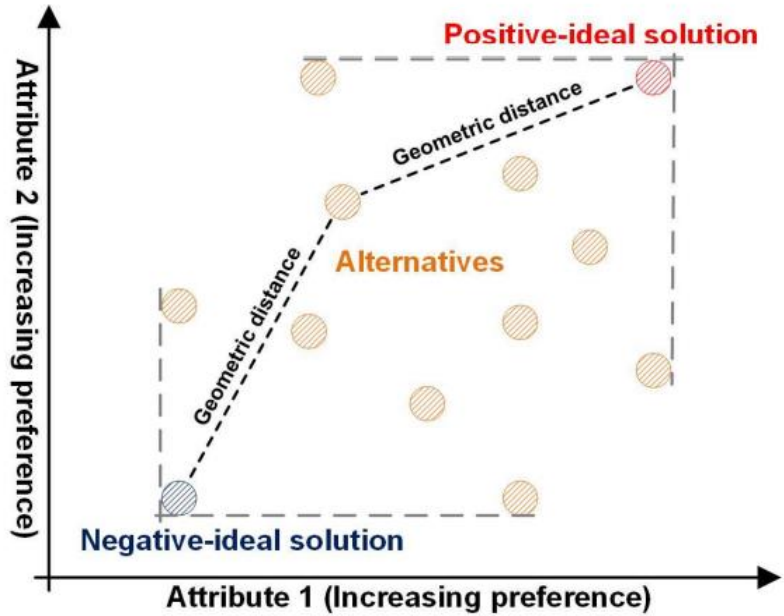

**Figure 2.** Graphical illustration of calculating the distance from the positive and negative ideal solution in TOPSIS (redrawn from Hwang and Yoon)

In this study, the application of TOPSIS method proceeds as follows: (1) normalising the values for each indicator, (2) identifying the positive-ideal solution (PIS) and negative-ideal solution (NIS) for each indicator, (3) calculating distances for each topic from PIS and NIS, and (4) prioritising the topics based on the distances. First, each attribute is converted to a unit vector and then multiplied by a predefined weight, as shown in Equation 1.

$$r_{ij} = \frac{w_j x_{ij}}{\sqrt{\sum_{k=1}^{|c_{T,i}|} x_{kj}}} \qquad \text{Equation 1}$$

Here, $x_{ij}$ is the original value of $i^{th}$ topic regarding $j^{th}$ emergence indicator, and $w_j$ is the weight of $j^{th}$ indicator, where $\sum w_j = 1$. Next, the vector normalisation is conducted to transform the scale of the indicators into a comparable form. The PIS and NIS, which have the best and worst values for each indicator, are identified. Here, it is necessary to define whether a large index value is a good alternative or the opposite, depending on the characteristics of the indicator. Based on previous studies on quantitative emergence indicators, we define that the higher the novelty, the lower the growth, coherence and community, the better the alternative. Accordingly, the PIS and NIS can be defined as follows:

$$PIS = [Max(Novelty), Min\ (Growth), Min\ (Coherence), Min\ (Community)]$$

$$NIS = [Min(Novelty), Max\ (Growth), Max\ (Coherence), Max\ (Community)]$$

Then, the distances from each topic to PIS and NIS are calculated using the Euclidean distance. Based on the distances from PIS ($d_i^+$) and NIS ($d_i^-$), the customer-stated topics can be prioritised with the integrated emergence index $P_i$.

$$P_i = \frac{d_i^-}{d_i^+ + d_i^-}$$

A high $P_i$ value of a topic means that the topic is far from NIS and close to PIS, which means that it has good index values overall. Consequently, customer-stated topics with high $P_i$ are highly likely to be emerging in the near future as they are highly evaluated in terms of novelty, growth, coherence and community. Therefore, the topics with high emergence score should be monitored, as they are expected to be of greater importance.

4. Case study: Electric vehicles

4.1. Overview

Identification of emerging customer-stated topics is an important task for business-to-consumer companies, as the engagement and influence of customers increases in the overall business. Therefore, in this study, a case study of the social media about the electronic vehicle (EV) is conducted for the following reasons. First, the EV industry is growing rapidly worldwide based on eco-friendly energy sources (Cohen, Lobel, & Perakis, 2016; Yuan & Li, 2021). Compared to the existing internal combustion automobile industry, the EV industry is newer, thus competition among EV companies is intensifying (Lin & Sovacool, 2020). In addition, as the number of consumers purchasing EVs increases rapidly, the needs of customers will become more diverse and complex. In addition, compared to other industries, since EV technology has not reached the maturity stage, it is highly likely that various issues or topics about products, functions, technology, and safety exist. Thus, identifying the customer-stated topics in the EV industry, tracking their changes, and uncovering emerging topics can provide companies with the intelligence regarding future opportunities. In this section, we tested the proposed approach on the social media data of Tesla, a leading EV manufacturer.

4.2. Data collection

The social media data collected for this analyses were online forum posts and comments. Although there are many social media platforms available for customers' opinions, we intend to obtain more diverse and ample data source that includes customers' expectation, interests, experience, complaints or discussion. To this end, we collect customer-generated data from Reddit, where online communities post, comment, discuss, and share opinions and information within subreddit that represents a specific topic. Specifically, we collected 2,816,285 posts and comments between January 1, 2015 and October 1, 2019 from the subreddit of Tesla (https://www.reddit.com/r/teslamotors/). This subreddit is not about a specific product or service, but about Tesla, a representative EV company. In fact, in that subreddit, not only EV products, but also various topics such as brand image of EV companies, comparison with competitors, opinions on eco-friendly energy sources, and information on EVs are being discussed. The quantitative trend of the collected data is shown in Figure 3. Overall, it is confirmed that the data is steadily increasing over time and that interest in Tesla vehicles (and electric vehicles in general) has steadily increased. In addition, we observed several peak points (reflected in Figure 3) on September 30, 2015, April 01, 2016, October 20, 2016, November 17, 2017, and March 15, 2019, and confirmed the events that led to these peaks—the release of Model X, the announcement of Model 3, and the announcement of the full self-driving function, Roadster, and Model Y, respectively.

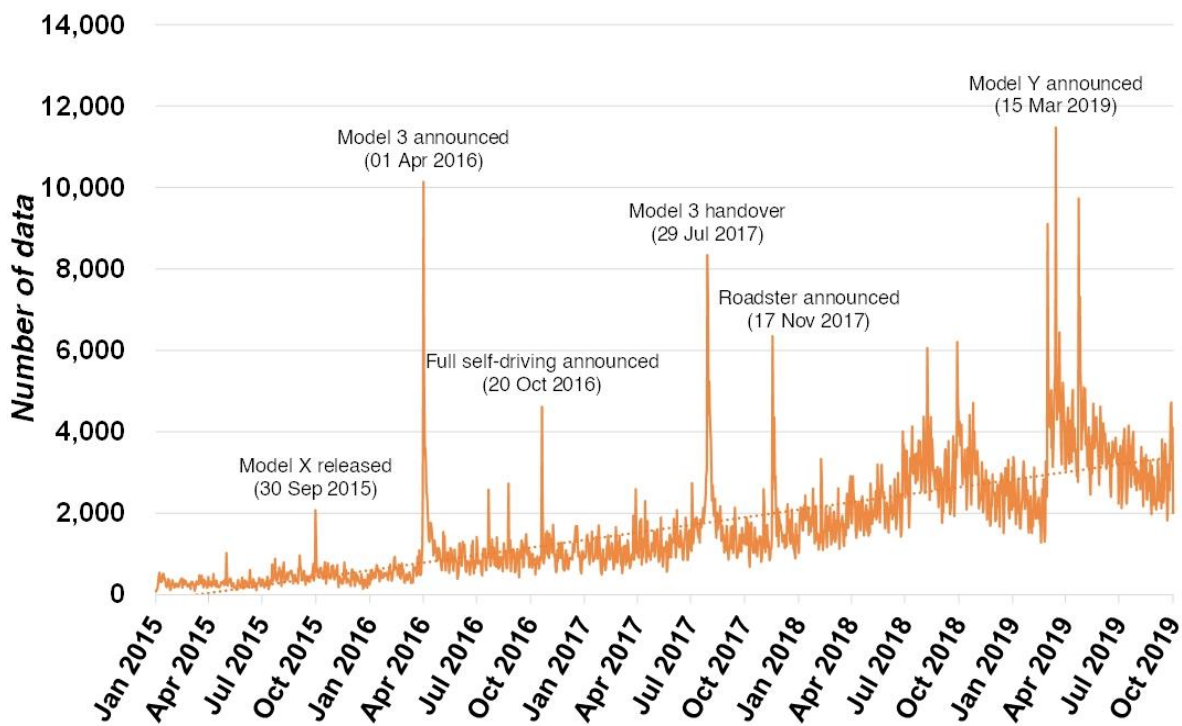

**Figure 3.** Quantitative trends of the collected data

4.3. Application of the automatic topic evolution detection method

4.3.1. Data pre-processing

Accurate identification of customer-stated topics requires securing high-quality dataset without noise. To this end, we conducted a pre-processing. First, noun chunks are extracted from documents by spaCy, a representative natural language processing python open library. The noun chunks can be nouns or noun phrases, which can indicate key parts of document or topic. Consequently, 844,878 noun chunks were extracted and stored. Second, the noun chunks were filtered as some of them do not represent customer-stated topics by themselves or may not be relevant to the main content of the case study. Therefore, we applied the rule-based filtering method and frequency-based method sequentially. The rule-based method utilises pre-defined rules regarding stopwords, length of the chunk, and regular expression. Specifically, we used the stopwords list provided by spaCy (Honnibal & Montani, 2017) and excluded the chunks that are too short to understand the meaning or too long to be misspelled. Here, we defined chunks with a length between 6 and 40 as valid, considering the mean and standard deviation, based on the length distribution of the chunks. In addition, based on the document frequency distribution, we excluded keywords with a frequency of 5 or less, which is the bottom 20%, from the analyses. They are rarely used by consumers, and even if important, other keywords with similar meanings will replace them. Consequently, the set consisting of 17,743 valid keywords ($V$) is obtained.

4.3.2. Topic evolution analyses

Thereafter, the set of filtered keywords was converted to an embedding matrix of filtered keywords by year, $V_d$, with a pre-trained SLM. Here, we used the RoBERTa based sentence-BERT, which is a successor of the BERT and designed to be more robust than the BERT by using extremely large parameters and data by Facebook AI Research (Liu, et al., 2019). Particularly, sentence-BERT (Reimers & Gurevych, 2019) provides an easy-to-use pre-trained model of RoBERTa. Since the embedding dimension is 1024 and we set the time slice to a year, the size of the embedding matrix ($V_d$) is 67878×1024. Table 4 shows the number of keywords and clusters by time slices. According to Algorithm 1, the optimal number of clusters was found for each time slice, and 138 topics were derived during the analyses period. Each topic is composed of similar keywords to indicate one topic, and labelling of a topic can be performed based on keywords (representative keywords) that are close to the centroid of the cluster. For example, the cluster (topic 19-6) with keywords such as 'tinted windows', 'brushed titanium', 'amber lights', 'black satin', 'glossy black', and 'invisible glass' at the centre is about the EV's external system. Also, the cluster, where the keywords such as 'cabin air', 'diesel emissions', 'global CO2 emissions', 'cabin air filter', 'emission regulations', and 'carbon emission' are main keywords, is the subject of air pollution. Likewise, the 138 topics labelled are summarised in Appendices 1.

**Table 4.** The number of keywords and clusters by time slice

| Time slice (*t*) | The number of keywords ($V_t$) | The number of clusters ($C_t$) |
| --- | --- | --- |
| 2015 | 8193 | 25 |
| 2016 | 12476 | 22 |
| 2017 | 14609 | 21 |
| 2018 | 16349 | 31 |
| 2019 | 16251 | 39 |

Next, we identified time-varying flows between customer-stated topics using Algorithm 2. The threshold ($\theta$) defining a valid relationship between topics was set to 0.1, the elbow point, based on the distribution of the number of topic links according to the threshold. In other words, topic pairs with less than 10% shared keywords are considered unrelated. The evolution of customer-stated topics can be visualised in Sankey diagram, as shown in Figure 4. The links between customer-stated topics can be interpreted by analysing the number of backward topics and forward topics. The topic with a single flow with both backward and forward topics can be regarded as constantly mentioned by many customers and the topics about cold climates (topic 17-10) belongs to this type. In the flow of the topic, the 'snow season', 'winter temperatures', 'colder climate', and 'subzero temperatures' appeared steadily as major keywords. This flow proceeded with the topics 15-23, 16-16, 17-10, 18-18 and 19-7 throughout the analyses period. The disadvantage of EVs is that their driving distance is shortened by a decrease in battery efficiency due to cold temperatures. Although various methods have been used to compensate for this problem, a complete solution is yet to be found. The flow shows that customers are also constantly interested in the 'cold temperature' problem. These topics need constant attention because they deal with consumer needs and problems that are constantly mentioned, and they are shown in Figure 4 as 'Constant topic'.

Topics with a single backward flow and multiple forward flows are categorised as Seed topics, which are those that have been continuously growing over time, but as the size of each topic becomes too large, they are differentiated by specific interests. For example, the topic 17-3 (Internal component), which includes 'internal space', 'glove compartment', 'black leather interior', and 'tight space' as major keywords, is representative. The initial topic regarding the internal components of EV was differentiated into the topics 18-26 (Interior), 18-2 (Vehicle control), 18-0 (Leather), and 18-3 (Hardware acceleration) in the next time slice. Therefore, the customers' interest in the internal component of EVs can be interpreted to have diversified into their design and technical elements. Thus, over time these four differentiated topics settled into Constant topics or were absorbed into other related ones. One the other hand, the topics with multiple backward flows and a single forward flow are named as Consolidate topics, which are composed of similar keywords that have existed in the periphery of extant topics. Therefore, these topics reflect an increase in customers' interests in the time slice, and thus we expect them to be coming into the spotlight. For example, the topic 18-3 (Hardware acceleration) is an example of consolidate topics, and its major keywords are 'onboard computers', 'physical space', 'virtual reality' and 'neural network'. A major content of this topic is "hardware acceleration for computer vision by using a dedicated chipset." The physical device-related key terms come from the topic 17-3 (Internal component), while the computer vision system-related key terms come from the topic 17-12 (Autopilot system). Our analyses shows that customer interest in 'hardware acceleration using dedicated chips' increased in 2017. In fact, Tesla announced that they are developing dedicated chips for autopilot at that time.

Lastly, topics that have multiple backward and forward flows are categorised as Temporary topics, which are likely to be temporarily combined topics because they contain similar contents and customers pay relatively less attention to them than other topics at that time. As shown on Figure 4, the temporary topics separated in subsequent time slices. For instance, topic 18-14 (Brakes & tires of a trailer) represents this type, with 'truck tires', 'trailer brakes', 'road tires', and 'disc brakes' as major key terms. The topic 18-14 (Brakes & tires of a trailer) may have been created by combining key terms from the topic 17-1 (Traffic control) and topic 17-15 (Steering control) in the prior time slice and separated into the topics 19-0 (Tire), 19-27 (Driving), and 19-33 (Brakes) in the next time slice.

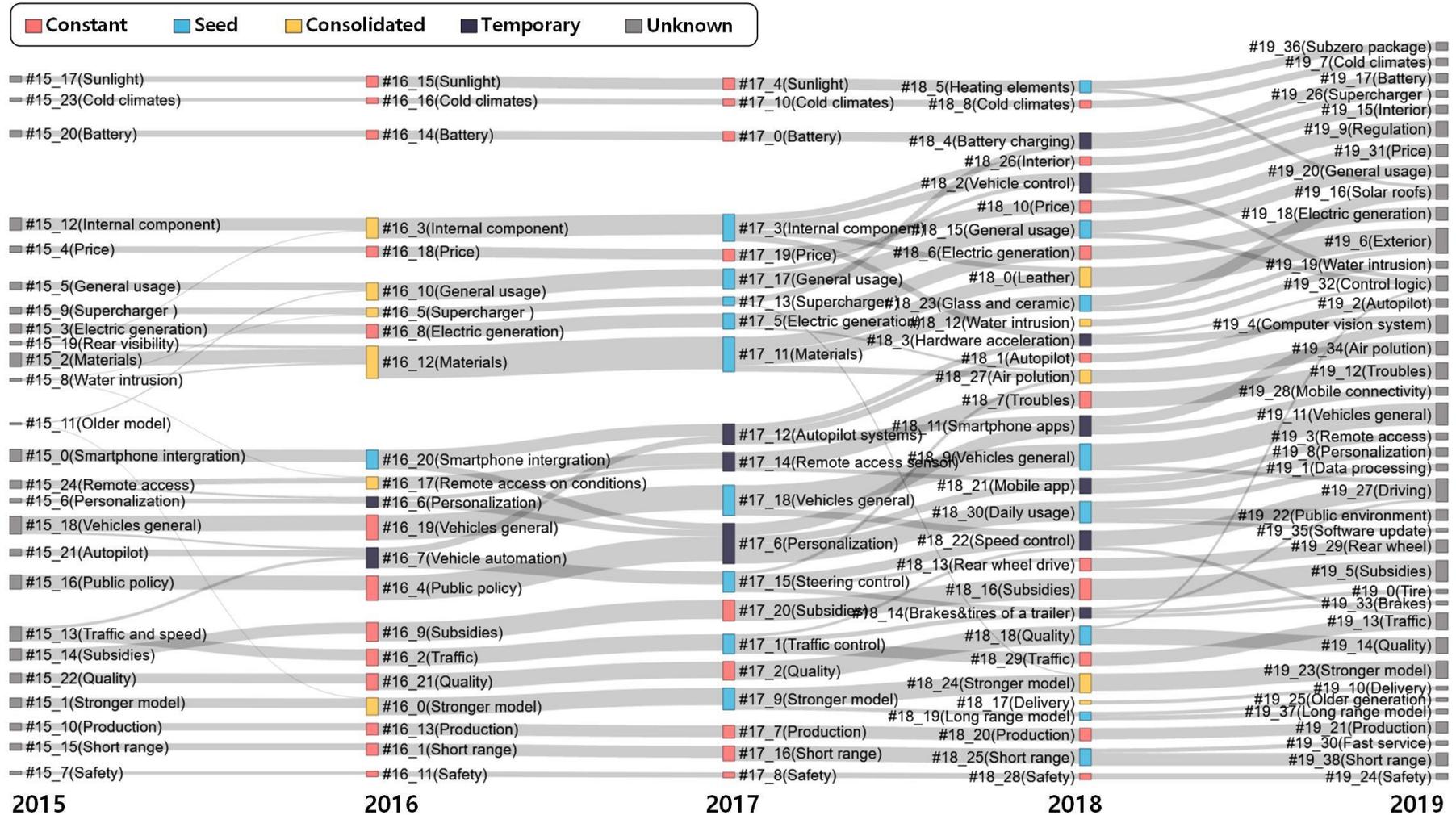

**Figure 4.** the evolution map of customer-stated topics from 2015 to 2019

### 4.3.3. Emergence scoring

Finally, we calculated the emergence attributes of 39 topics in 2019 to identify emerging customer-stated topics, which are expected to be of greater importance after 2019. The emergence indicator values of topics are incorporated into an integrated emergence score via TOPSIS. The weights of each attribute were equally assigned, meaning that all attributes have identical importance in this case study. Table 5 shows the top 10 emerging topics in 2019.

**Table 5.** Top 10 emerging customer-stated topics in the last time slice

| Customer-stated topics (Identification no.-Label) | Novelty | Growth | Coherence | Community | Emergence |
|---|---|---|---|---|---|
| Topic.19-6 (Exterior) | 0.0801 | 0.0388 | 0.0212 | 0.0309 | 0.8153 |
| Topic. 19-19 (Regulation) | 0.0543 | 0.0393 | 0.0272 | 0.0322 | 0.6732 |
| Topic.19-4 (Computer vision systems) | 0.0600 | 0.0444 | 0.0294 | 0.0347 | 0.6686 |
| Topic.19-32 (Control logic) | 0.0520 | 0.0329 | 0.0361 | 0.0290 | 0.6675 |
| Topic.19-12 (Troubles) | 0.0477 | 0.0367 | 0.0279 | 0.0240 | 0.6594 |
| Topic.19-5 (Subsidies) | 0.0589 | 0.0402 | 0.0400 | 0.0382 | 0.6333 |
| Topic.19-34 (Air pollution) | 0.0479 | 0.0377 | 0.0252 | 0.0372 | 0.6229 |
| Topic.19-27 (Driving) | 0.0651 | 0.0388 | 0.0405 | 0.0521 | 0.6066 |
| Topic.19-22 (Public environment) | 0.0419 | 0.0352 | 0.0297 | 0.0310 | 0.6018 |
| Topic.19-1 (Data processing) | 0.0436 | 0.0378 | 0.0327 | 0.0333 | 0.5869 |
| Average | 0.0366 | 0.0395 | 0.0391 | 0.0383 | 0.4993 |

Finally, emerging topics were prioritised via TOPSIS, and parts of them are listed in Table 6. These topics are likely to be important in the near future after 2019, considering the emergence index values in 2019. First, there are topics about the advanced technologies that underlies EVs, such as topics 19-1 (Data processing), 19-4 (Computer vision system), 19-27 (Driving) and 19-32 (Control logic). These topics are highly likely to be emerging as customers' understanding and expectation of autopilot function increase. This means that autopilot and its related functions will occupy a significant portion of customers' interests soon. Also, interestingly, Tesla's customers are concerned about government regulations. For instances, we identified topics that are either directly or indirectly related to government policies and regulations including topics 19-5 (Subsidies), 19-9 (Regulation), and 19-22 (Public environment). Other topics such as 19-9 (Regulation) and 19-22 (Public environment), are related to the US governments' regulations for fuel (electricity) consumption, carbon emissions, and mandatory sales. Due to these regulations, Tesla limits motor performance and autopilot function. In addition, it seems that customers are also interested in existing regional differences and the regulations that exist in their residential areas. In some cases, even if it is not government regulation, Tesla has restrictions based on safety or technical reasons. Such topics are reflected in terms such as "Maximum plaid," which appeared as a major key term in the topic 19-9 (regulation). This keyword refers to a mode that can release the technically limited acceleration performance feature in Tesla EVs, showing that many customers are interested in high driving performance. Topic 19-5 (Subsidies) is related to subsidies paid to electric car buyers in the US and has appeared steadily as topics 15-14, 16-9, 17-20, 18-16, and 19-5. Since these topics appear across the time slices, they fall under Constant topics. Since the topic 19-5 (Subsidies) is emerging, it can be predicted that interest in subsidies will increase, thus Tesla should act by developing a cheaper EV model.

**Table 6.** Main keywords and backward topics of top 10 emerging topics

| Ranking | Customer-stated topics | Label | Major keywords | Backward topics (Label) |
|---|---|---|---|---|
| 1 | 19-6 | Exterior | Tinted windows, Brushed titanium, Amber lights, Black satin, Glossy black, Invisible glass | 18-0 (Leather), 18-23 (Glass) |
| 2 | 19-9 | Regulation | Time limit, Legal approval, Legal agreements, State regulations, Maximum plaid, Proper permits | 18-2 (Car control) |
| 3 | 19-4 | Computer vision systems | Computer vision, Onboard computers, Computer control, Virtual reality, Computer graphics, Computer power | 18-3 (Hardware acceleration), 18-11 (Mobile app) |
| 4 | 19-32 | Control logic | Solid logic, Clear markings, Control logic, Perfect response, Proper control, Extreme precision | 18-2 (Car control), 18-3 (Hardware acceleration), 18-15 (General usage), 18-18 (Quality) |
| 5 | 19-12 | Troubles | Problem areas, Obvious flaws, Error bars, Obvious defects, Wrong link | 18-7 (Troubles) |
| 6 | 19-5 | Subsidies | Taxpayer subsidies, Payment method, Monetary compensation, Pricing structure, Price savings | 18-16 (Subsidies) |
| 7 | 19-34 | Air pollution | Cabin air, Diesel emissions, Global $CO_2$ emissions, Cabin air filter, Emission regulations, Carbon emission | 18-27 (Air pollution) |
| 8 | 19-27 | Driving | Driving functions, Vehicle speed, Vehicle power, Interstate driving, Driving data, Driving mode | 18-9 (Vehicle general), 18-22 (Speed control), 18-14 (Brakes & Tires of trucks) |
| 9 | 19-22 | Public environment | Public space, Urban transport, Metropolitan areas, Public policy, Public roadways, Urban roads | 18-30 (Daily usage) |
| 10 | 19-1 | Data processing | Visual processing, Data transmission, Video uploads, Data capture, Data connections, | 18-30 (Daily usage) |

Performance data

5. Conclusion

This study proposes a novel approach to detect the dynamics of customer-stated topics and identify emerging topics, thereby fully automating the customer language understanding process. The main goal of the proposed approach can be summarised as understanding the dynamic changes of the topics and their possibilities as future opportunities. We design two functions of tracking the dynamics of topics, and anticipating the future of topics with quantitative emergence scores. This automatic approach requires limited expert intervention and is robust against changes, updates, and additions of the initial dataset. In this study, the case study was conducted on Tesla products, a leading EV manufacturer. We drew some interesting interpretations for both the past and the near future. For example, there are four types of topics depending on the number of backward and forward flows of interest: the constant, consolidated, seed, and temporary. There are five constant flows of topics such as cold climate, regenerative braking, quality of the vehicle, safety, and production. There are the topics that are likely to emerge, some of which are related to the exterior features of Tesla cars are the most emerging, or related to autopilot features. In addition, the case study identified four customer needs regarding exterior, autopilot, performance, and price. The case study showed that the proposed approach can analyse past flows and anticipate emerging topics based on voluminous social media data.

We expect that this study will make both academic and industrial contributions to relevant fields. First, the approach proposed is meaningful as it addresses emerging topics with topic evolution analyses, which includes not only interpreting historical data but also envisioning the near future. Because customer-stated topics are dynamic, it is important to understand the dynamics of topics during the analyses period. Second, this study is the initial attempt to define and measure emerging attributes of customer-stated topics in our best knowledge. As prior studies related to the attributes of an emerging entity have focused on technology or research, this study could be a milestone in emerging customer-stated topic studies, extending the use of emerging attributes to anticipating of customer needs. From the perspective of practice, the proposed approach and quantitative outcomes can be used for various customer-centric business tasks because constructing a unique relationship with both existing and new customers is important for the sustainable growth. For example, in marketing where the selection of an appropriate sales point for a product or service is crucial, the proposed approach guides experts to identify rapidly rising needs and establishing timely marketing strategies. The results can be used as objective material to understand the functions that customers want and determine how these needs will be incorporated into new products or functions.

# Appendices

**Appendices 1.** List of 138 customer-stated topics

| Customer-stated topic No. | Label | CI# | Label | CI# | Label |
|---|---|---|---|---|---|
| Topic.15-0 | Smartphone integration | Topic.16-21 | Quality | Topic.18-24 | Stronger model |
| Topic.15-1 | Stronger model | Topic.17-0 | Battery | Topic.18-25 | Short range |
| Topic.15-2 | Materials | Topic.17-1 | Traffic control | Topic.18-26 | Interior |
| Topic.15-3 | Electric generation | Topic.17-2 | Quality | Topic.18-27 | Air pollution |
| Topic.15-4 | Price | Topic.17-3 | Internal component | Topic.18-28 | Safety |
| Topic.15-5 | General usage | Topic.17-4 | Sunlight | Topic.18-29 | Traffic |
| Topic.15-6 | Personalisation | Topic.17-5 | Electric generation | Topic.18-30 | Daily usage |
| Topic.15-7 | Safety | Topic.17-6 | Personalisation | Topic.19-0 | Tire |
| Topic.15-8 | Water intrusion | Topic.17-7 | Production | Topic.19-1 | Data processing |
| Topic.15-9 | Supercharger | Topic.17-8 | Safety | Topic.19-2 | Autopilot |
| Topic.15-10 | Production | Topic.17-9 | Stronger model | Topic.19-3 | Remote access |
| Topic.15-11 | Older model | Topic.17-10 | Cold climates | Topic.19-4 | Computer vision system |
| Topic.15-12 | Internal component | Topic.17-11 | Materials | Topic.19-5 | Subsidies |
| Topic.15-13 | Traffic and speed | Topic.17-12 | Autopilot systems | Topic.19-6 | Exterior |
| Topic.15-14 | Subsidies | Topic.17-13 | Supercharger | Topic.19-7 | Cold climates |
| Topic.15-15 | Short range | Topic.17-14 | Remote access sensor | Topic.19-8 | Personalisation |
| Topic.15-16 | Public policy | Topic.17-15 | Steering control | Topic.19-9 | Regulation |
| Topic.15-17 | Sunlight | Topic.17-16 | Short range | Topic.19-10 | Delivery |
| Topic.15-18 | Vehicles general | Topic.17-17 | General usage | Topic.19-11 | Vehicles general |
| Topic.15-19 | Rear visibility | Topic.17-18 | Vehicles general | Topic.19-12 | Troubles |
| Topic.15-20 | Battery | Topic.17-19 | Price | Topic.19-13 | Traffic |
| Topic.15-21 | Autopilot | Topic.17-20 | Subsidies | Topic.19-14 | Quality |
| Topic.15-22 | Quality | Topic.18-0 | Leather | Topic.19-15 | Interior |
| Topic.15-23 | Cold climates | Topic.18-1 | Autopilot | Topic.19-16 | Solar roofs |
| Topic.15-24 | Remote access | Topic.18-2 | Vehicle control | Topic.19-17 | Battery |
| Topic.16-0 | Stronger model | Topic.18-3 | Hardware acceleration | Topic.19-18 | Electric generation |
| Topic.16-1 | Short range | Topic.18-4 | Battery charging | Topic.19-19 | Water intrusion |
| Topic.16-2 | Traffic | Topic.18-5 | Heating elements | Topic.19-20 | General usage |
| Topic.16-3 | Internal component | Topic.18-6 | Electric generation | Topic.19-21 | Production |
| Topic.16-4 | Public policy | Topic.18-7 | Troubles | Topic.19-22 | Public environment |
| Topic.16-5 | Supercharger | Topic.18-8 | Cold climates | Topic.19-23 | Stronger model |
| Topic.16-6 | Personalisation | Topic.18-9 | Vehicles general | Topic.19-24 | Safety |
| Topic.16-7 | Vehicle automation | Topic.18-10 | Price | Topic.19-25 | Older generation |
| Topic.16-8 | Electric generation | Topic.18-11 | Smartphone apps | Topic.19-26 | Supercharger |
| Topic.16-9 | Subsidies | Topic.18-12 | Water intrusion | Topic.19-27 | Driving |
| Topic.16-10 | General usage | Topic.18-13 | Rear wheel drive | Topic.19-28 | Mobile connectivity |
| Topic.16-11 | Safety | Topic.18-14 | Brakes & tires of a trailer | Topic.19-29 | Rear-wheel drive |
| Topic.16-12 | Materials | Topic.18-15 | General usage | Topic.19-30 | Fast service |
| Topic.16-13 | Production | Topic.18-16 | Subsidies | Topic.19-31 | Price |
| Topic.16-14 | Battery | Topic.18-17 | Delivery | Topic.19-32 | Control logic |
| Topic.16-15 | Sunlight | Topic.18-18 | Quality | Topic.19-33 | Brakes |
| Topic.16-16 | Cold climates | Topic.18-19 | Long range model | Topic.19-34 | Air pollution |
| Topic.16-17 | Remote access on conditions | Topic.18-20 | Production | Topic.19-35 | Software update |

| Topic.16-18 | Price | Topic.18-21 | Mobile app | Topic.19-36 | Subzero package |
| --- | --- | --- | --- | --- | --- |
| Topic.16-19 | Vehicles general | Topic.18-22 | Speed control | Topic.19-37 | Long range model |
| Topic.16-20 | Smartphone integration | Topic.18-23 | Glass and ceramic | Topic.19-38 | Short range |